\documentclass[11pt]{article}
\usepackage{amsmath}
\usepackage[preprint]{acl}

\usepackage{times}
\usepackage{latexsym}
\usepackage{amsfonts}
\usepackage{booktabs}
\usepackage[T1]{fontenc}
\usepackage{amssymb,enumitem}
\usepackage{xcolor}

\usepackage[utf8]{inputenc}
\usepackage{url}
\usepackage{microtype}
\usepackage{enumitem}
\usepackage{multirow} 
\usepackage{inconsolata}
\usepackage{tabularx}
\usepackage{graphicx}

\usepackage{multirow}
\usepackage{graphicx}
\usepackage[table,xcdraw,dvipsnames]{xcolor}
\usepackage{arydshln}

\title{Mixture-of-Experts with Intermediate CTC Supervision \\ for Accented Speech Recognition}

\author{Wonjun Lee, Hyunghun Kim, Gary Guenbae Lee \\
        Address line \\ ... \\ Address line}

\author{Wonjun Lee$^{\lozenge}$, Hyounghun Kim$^{\lozenge, \blacklozenge}$ \and Gary Geunbae Lee$^{\lozenge, \blacklozenge}$ \\
     $^{\lozenge}$  Department of Computer Science and Engineering, POSTECH \\     
	 $^{\blacklozenge}$Graduate School of Artificial Intelligence, POSTECH \\ 
    \texttt{\{lee1jun, h.kim, gblee\}@postech.ac.kr}
    }

\begin{document}
\maketitle
\begin{abstract}
Accented speech remains a persistent challenge for automatic speech recognition (ASR), as most models are trained on data dominated by a few high-resource English varieties, leading to substantial performance degradation for other accents. 
Accent-agnostic approaches improve robustness yet struggle with heavily accented or unseen varieties, while accent-specific methods rely on limited and often noisy labels. 
We introduce \textsc{Moe-Ctc}, a Mixture-of-Experts architecture with intermediate CTC supervision that jointly promotes expert specialization and generalization. 
During training, accent-aware routing encourages experts to capture accent-specific patterns, which gradually transitions to label-free routing for inference. 
Each expert is equipped with its own CTC head to align routing with transcription quality, and a routing-augmented loss further stabilizes optimization. 
Experiments on the \textsc{Mcv-Accent} benchmark demonstrate consistent gains across both seen and unseen accents in low- and high-resource conditions, achieving up to 29.3\% relative WER reduction over strong FastConformer baselines.
\end{abstract}

\section{Introduction}

Automatic speech recognition (ASR) has reached near-human accuracy on benchmarks such as LibriSpeech~\cite{librispeech} and Switchboard, driven by large-scale pretraining and advanced architectures~\cite{baevski2020wav2vec, fastconformer}.  
However, these advances do not generalize uniformly: performance drops markedly on accented speech, where acoustic and phonetic patterns differ from those dominant in training data.  
Because accented speech remains underrepresented in most corpora, current ASR systems under-perform for accented speakers, raising concerns of fairness and inclusivity~\cite{shor2019personalizing, Dheram2022toward, graham2024whisper_accents}.

Prior research has mainly followed two directions.  
\textbf{Accent-agnostic approaches} aim to learn robust, accent-invariant representations through self-supervised or adversarial objectives~\cite{hsu2021hubert, das2021adversarial}.  
While models such as XLS-R \cite{babu2022xls} and Whisper \cite{radford2023robust} improve general robustness, their performance still degrades for heavily accented or unseen varieties, indicating that full invariance is insufficient to capture accent diversity~\cite{zuluaga2023commonaccent}.  

\textbf{Accent-specific approaches}, on the other hand, leverage accent labels for explicit specialization via fine-tuning~\cite{li2021accentrobustautomaticspeechrecognition}, data augmentation~\cite{do2024improving}, or parameter-efficient modules such as adapters and accent embeddings~\cite{tomanek-etal-2021-residual, jain2018accentemb}.  
These methods improve performance for known accents but depend on accent labels even at inference, limiting scalability and generalization to unseen speakers.

A scalable alternative lies in \textbf{Mixture-of-Experts (MoE)}, which allows experts to specialize across domains.  
Since accent is an utterance-level attribute, sequence-level MoE provides a natural fit for accented ASR. 
Recent studies have explored MoE for multilingual or dialectal ASR~\cite{you2021speechmoe, zhou2024dialectmoe, bagat2025mixture}, yet existing methods either require accent labels at test time or fail to reliably route inputs to the appropriate experts.

To address these challenges, we propose \textsc{Moe-Ctc}, a novel architecture that integrates MoE with expert-level intermediate CTC (Connectionist Temporal Classification) supervision.  
Our method learns accent-aware routing during training but transitions to accent-agnostic inference through a two-stage learning process.  
Each expert is equipped with an auxiliary CTC head, aligning routing decisions with transcription quality and improving optimization stability.  

Experiments on the \textsc{Mcv-Accent} benchmark~\cite{codebook2023} demonstrate that \textsc{Moe-Ctc} consistently outperforms prior accent-robust baselines, achieving substantial WER reductions across both seen and unseen accents.

\section{Related Work}

\subsection{Accented Speech Recognition}
Accented speech has long been recognized as a major source of performance degradation in ASR systems. Analyses on corpora such as L2-ARCTIC \cite{l2arctic}, CommonVoice \cite{commonvoice}, and GLOBE \cite{globe} consistently show that models trained primarily on native speech exhibit substantially higher error rates for accented speakers \cite{zuluaga2023commonaccent, shor2019personalizing}. Even large-scale pretrained models such as XLS-R and Whisper reduce but do not eliminate the accent gap: Whisper shows higher WER on non-native accents \cite{graham2024whisper_accents}, while XLS-R/Whisper performance degrades on African-accented datasets \cite{olatunji2023afrispeech200}.  

\noindent\textbf{Accent-agnostic Approaches.}  
One research strand aims to improve robustness without relying on accent labels. Self-supervised pretraining with wav2vec~2.0 \cite{baevski2020wav2vec}, HuBERT \cite{hsu2021hubert}, and related models has proven effective at learning accent-invariant representations. Similarly, large-scale multilingual pretraining (e.g., XLS-R, Whisper) improves cross-accent generalization. Other works adopt adversarial or domain generalization techniques—such as domain adversarial training \cite{sun2018dat, das2021adversarial}, relabeling strategies \cite{hu2020reDAT}, or coupled/contrastive learning \cite{unni2020coupled}—to enforce invariance. While these methods improve robustness overall, they still under-perform on heavily accented or unseen varieties compared to accent-aware models.  

\noindent\textbf{Accent-specific Approaches.}  
Another line of work explicitly incorporates accent supervision. Fine-tuning on accent-labeled corpora \cite{li2021accentrobustautomaticspeechrecognition} often boosts performance for the target accent but compromises generalization to unseen varieties. Data augmentation synthesizes accented speech via phoneme perturbations or accent conversion \cite{do2024improving}, though gains are limited by coverage and realism. Parameter-efficient adaptations such as residual adapters \cite{tomanek-etal-2021-residual}, non-linear modules \cite{qian2022layer}, and accent embeddings \cite{jain2018accentemb, chen2015ivector, li2017multidialect, viglino2019fusion} offer lighter-weight alternatives. Rapid cross-accent adaptation has also been explored for low-resource settings \cite{rao2017hierarchical, winata2020fastadapt}. Recently, \citet{bagat2025mixture} organized accent-specific LoRA modules as a mixture of experts, showing that modular parameterization can balance specialization and scalability.  

\subsection{Mixture-of-Experts for Accent Adaptation}
While accent-specific methods achieve strong specialization, they depend on labeled accent data and generalize poorly to unseen accents. MoE offers a scalable alternative: by selectively activating specialized subnetworks, MoE increases model capacity without proportional computational cost \cite{shazeer2017outrageously, fedus2022switch}. In ASR, MoE has been applied to multilingual and multi-domain recognition \cite{you2021speechmoe, hu2023mixture}, where routing mechanisms distribute inputs across experts under diverse acoustic conditions.  

A key challenge for accent-aware MoE is routing when accent labels are unavailable at inference. To address this, \citet{codebook2023, codebook2024} introduced beam search–based expert (codebook) selection, while \citet{zhou2024dialectmoe} proposed dialect-adaptive dynamic routing using feature–embedding combinations. \citet{bagat2025mixture} presented MAS-LoRA, where accent-specific LoRA modules act as experts; averages all LoRA experts equally when accent labels are unavailable.  

In contrast, our \textsc{Moe-Ctc} introduces accent-aware routing during early training to encourage expert specialization, then transitions to accent-agnostic training for inference. This avoids auxiliary routing strategies, requires no accent labels at test time, and enables effective generalization to unseen accents.

\section{Background}

Our proposed framework, \textsc{Moe-Ctc}, is built upon a FastConformer \cite{fastconformer} encoder 
 with the CTC head for ASR, augmented with two key ideas: 
\emph{intermediate CTC supervision} and the \emph{MoE} architecture. 
Intermediate CTC provides auxiliary objectives that stabilize optimization and enhance 
representation learning, while MoE introduces capacity through expert specialization.
In the following subsections, we review these two components to clarify the foundation of our approach.

\subsection{Intermediate CTC Loss}
\label{sec:intermediate-ctc}
Connectionist Temporal Classification (CTC) \cite{ctc} is a widely adopted training criterion for non-autoregressive ASR models. 
By marginalizing over all valid frame-to-label alignments, CTC enables end-to-end training without the need for frame-level annotations.  

Recent studies have shown that inserting intermediate CTC losses \cite{interctc} at multiple encoder layers can improve training stability and accelerate convergence \cite{komatsu2022better, hojo2024boosting}. 
These auxiliary objectives encourage hidden representations at lower layers to be predictive of output units, thereby mitigating vanishing gradients and facilitating optimization.  

Formally, let $h^{(\ell)}$ denote the encoder states at layer $\ell$.  
An auxiliary CTC loss $\mathcal{L}_{\text{CTC}}^{(\ell)}$ is computed, and the overall training objective is
\begin{equation}
\mathcal{L} = \mathcal{L}_{\text{CTC}}^{(L)} + \sum_{\ell \in \mathcal{I}} \lambda_\ell \, \mathcal{L}_{\text{CTC}}^{(\ell)},
\label{eq:interctc}
\end{equation}
where $L$ is the final encoder layer, $\mathcal{I}$ is the set of intermediate layers with auxiliary supervision, and $\lambda_\ell$ are weighting coefficients.  


\subsection{Sequence-level Mixture-of-Experts}

Mixture-of-Experts (MoE) \cite{shazeer2017outrageously, fedus2022switch} augments deep models with sparsely activated expert networks. 
A gating function routes inputs to a subset of experts, enabling large capacity without proportional computation.  

Conventional MoE performs token-level routing, where each audio frame in ASR may be directed to different experts, incurring high computation and switching overhead.  
In contrast, sequence-level MoE assigns an entire utterance to one or a few experts, reducing cost while preserving specialization.  
This is particularly suitable for accented speech, since accent is an utterance-level attribute.  

Given encoder input $H \in \mathbb{R}^{B \times T \times D}$, where $B$ is the batch size, $T$ the number of audio frames, and $D$ the hidden dimension, we mean-pool across time:
\begin{equation}
\bar{h} = \tfrac{1}{T} \sum_{t=1}^{T} H_{:,t,:}.
\label{eq:pool}
\end{equation}
The routing network maps this pooled representation $\bar{h} \in \mathbb{R}^{B \times D}$ to logits 
$L \in \mathbb{R}^{B \times N}$, where $N$ is the number of experts and $L_{i,j}$ is the logit for assigning sample $i$ to expert $j$.  
The gating probabilities are then obtained via softmax:
\[
g_{i,j} = \frac{\exp(L_{i,j})}{\sum_{k=1}^{N}\exp(L_{i,k})}
\]
Finally, the MoE output for sample $i$ is the weighted sum of expert outputs:
\[
\textsc{MoE}(H_i) = \sum_{j=1}^{N} g_{i,j}\,\textsc{Expert}_j(H_i).
\]

Here, each $\textsc{Expert}_j$ is implemented as a feed-forward block, 
which we describe in detail in the following subsection.

\section{Method}

\begin{figure*}
    \centering
    \includegraphics[width=0.90\linewidth]{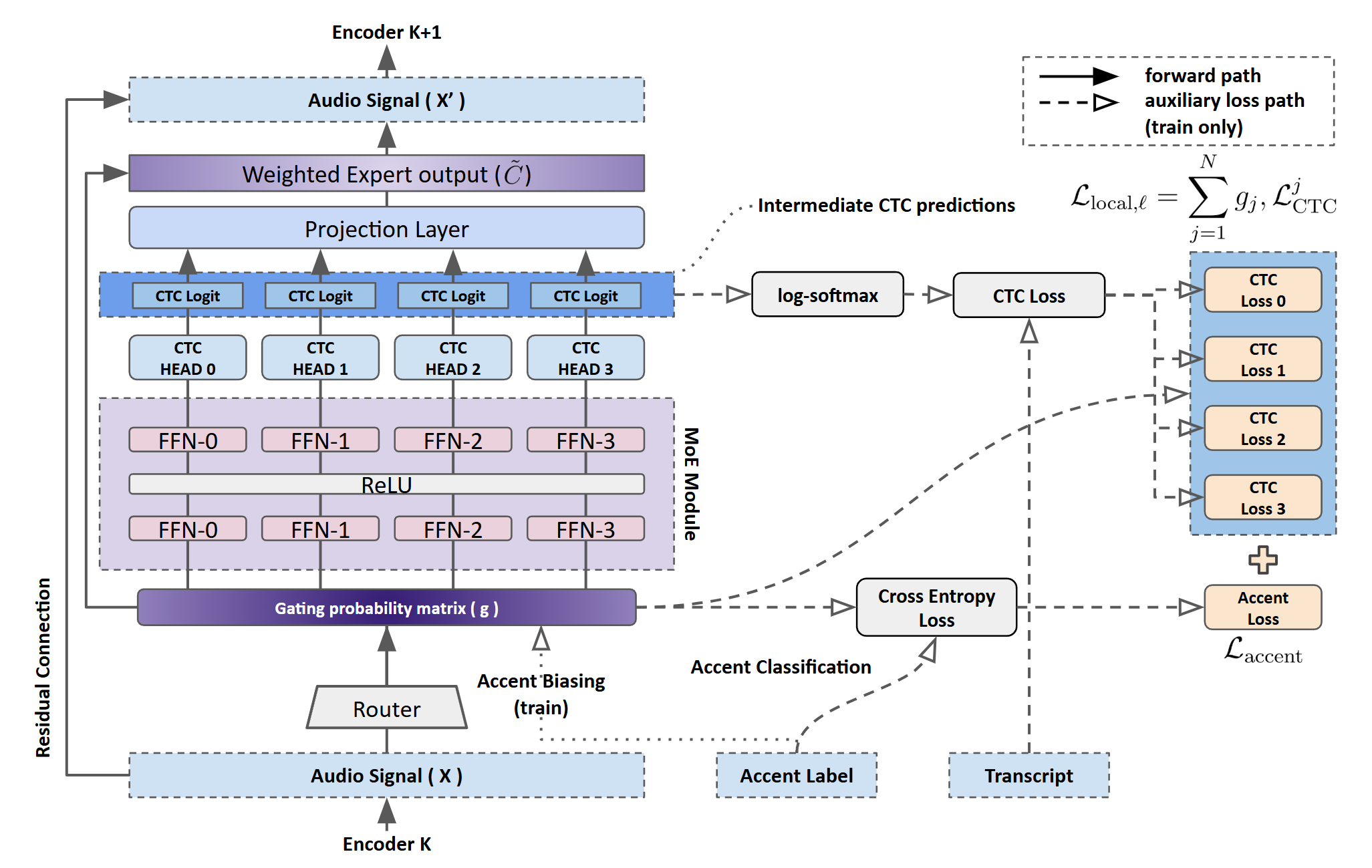}
    \caption{Overview of the proposed \textsc{Moe-Ctc} architecture. 
    The $\ell$-th MoE module, inserted between encoder blocks, is illustrated with four experts (shown as an example). 
    Each expert is equipped with an auxiliary CTC head, producing local supervision loss $\mathcal{L}_{\text{local},\ell}$. 
    During training, the router incorporates accent-aware routing through Accent Biasing and Accent Classification loss ($\mathcal{L}_{\text{accent}}$).}

    \label{fig:moectc}
\end{figure*}

\subsection{Mixture-of-Experts Module}

We extend the FastConformer encoder by inserting MoE modules between encoder layers ($\ell$-th), rather than replacing the internal feed-forward networks as in \citet{hu2023mixture}.  
Each MoE module consists of $N$ parallel experts, where each expert is a two-layer feed-forward block with ReLU activation:
\[
\textsc{Expert}_j(H) = \textsc{FFN}_{2,j}\big(\text{ReLU}(\textsc{FFN}_{1,j}(H))\big),
\]
with $\textsc{FFN}_{1,j}, \textsc{FFN}_{2,j} \in \mathbb{R}^{D \times D}$ and $D$ the hidden dimension.  

The routing network maps the pooled encoder representation $\bar{h}_i$ of sample $i$ (Eq.~\ref{eq:pool}) 
to a distribution $g_i \in \mathbb{R}^{N}$ over $N$ experts.  
To promote efficiency, we adopt top-$K$ routing: only the $K$ experts with the highest probabilities $g_{i,j}$ are selected, and their weights are renormalized:
\[
\tilde{g}_{i,j} = \frac{g_{i,j}}{\sum_{k \in \mathcal{K}_i} g_{i,k}}, \quad j \in \mathcal{K}_i,
\]
where $\mathcal{K}_i \subseteq \{1,\dots,N\}$ denotes the selected top-$K$ experts for sample $i$.  
The final MoE output for sample $i$ is then the weighted combination:
\[
\textsc{Moe}(H_i) = \sum_{j \in \mathcal{K}_i} \tilde{g}_{i,j}\,\textsc{Expert}_j(H_i).
\]

\subsection{Accent-Aware Routing}
\label{sec:accent-aware}

Building on the generic MoE module, we introduce \textbf{\textsc{Accent-MoE}}, which incorporates explicit accent supervision.  
During training, each utterance is associated with an accent label $a_i \in \{0,1,\dots,A-1\}$. 
We align each accent to a designated expert, encouraging balanced utilization and accent-specific specialization.  
For example, align $j$-th expert for the Scotland accent ($a_i)$.

Formally, let $L_{i,j}$ denote the gating logit for sample $i$ and expert $j$, and $a_i$ its accent index.  
We define the accent-biased logit:
\begin{equation}
\tilde{L}_{i,j} = L_{i,j} + \alpha \cdot \mathbf{1}[j=a_i], 
\label{eq:bias}
\end{equation}
where \textbf{$\alpha$} is the bias strength and $\mathbf{1}[\cdot]$ is the indicator function.  
The final routing distribution is then
\begin{equation*}
g_{i,j} = \frac{\exp(\tilde{L}_{i,j})}{\sum_{k=0}^{N-1}\exp(\tilde{L}_{i,k})}. 
\end{equation*}

\vspace{0.5em}


\noindent To enable the router to select appropriate experts even without accent labels at test time, 
the gating weights are also interpreted as accent logits and trained with an auxiliary loss
\begin{equation}
\mathcal{L}_{\text{accent}} = -\sum_i \log \frac{\exp(g_{i,a_i})}{\sum_{j=0}^{N-1} \exp(g_{i,j})}.
\end{equation}
By applying the biasing term, we guide the model to route samples to accent-specific experts—so that each expert is primarily trained for its designated accent—while the auxiliary classification loss further regularizes the routing process.

However, explicit accent supervision alone may not align routing with recognition quality, so we introduce expert-level CTC supervision to directly couple expert selection with transcription accuracy.

\subsection{Expert-Level CTC Supervision (\textsc{Moe-Ctc})}

Inspired by intermediate CTC loss (Section~\ref{sec:intermediate-ctc}), 
we extend this principle to \textsc{Accent-MoE} by equipping each expert with its own auxiliary CTC head, 
as illustrated in Figure~\ref{fig:moectc}. 
This design provides fine-grained supervision that guides both layer-wise representations and expert-level specialization, a strategy we denote as \textbf{\textsc{MoE-CTC}}.

Let $X \in \mathbb{R}^{B \times T \times D}$ denote the input to the MoE module, 
and $\textsc{Expert}_j(X)$ the output of the $j$-th expert.  
Each expert is equipped with a CTC head that maps hidden states into the vocabulary space:
\[
\textsc{Ctc-Head}: \mathbb{R}^{B \times T \times D} \rightarrow \mathbb{R}^{B \times T \times V},
\]
where $V$ is the vocabulary size of \textsc{Ctc-Head}.  
Each expert-specific head yields an auxiliary CTC loss 
$\mathcal{L}_{\text{CTC}}^{(\ell,j)}$ at layer $\ell$ for expert $j$, 
evaluated against the ground-truth transcription.  

Following the intermediate CTC self-conditioning strategy of \citet{scctc}, the logits from each expert’s CTC head are projected back into the hidden dimension through a learnable linear mapping,
\[
\textsc{Proj}: \mathbb{R}^{B \times T \times V} \rightarrow \mathbb{R}^{B \times T \times D},
\]
then combined according to the gating weights and added via a residual pathway:
\begin{align*}
H_j &= \textsc{Expert}_j(X), \\
C_j &= \textsc{Ctc-Head}_j(H_j), \\
P_j &= \textsc{Proj}(C_j), \\
\tilde{C} &= \sum_{j=1}^N g_{j}\,P_j \in \mathbb{R}^{B \times T \times D}, \\
X' &= X + \tilde{C}. \label{eq:residual}
\end{align*}

Here, the auxiliary CTC losses are computed directly from $C_j$, while the residual pathway ensures that these CTC-informed signals are stably integrated back into the shared representation.  
In this way, expert behavior is aligned with transcription quality and CTC-informed feedback is propagated across layers.  

Finally, to further couple expert performance with routing, we define a local routing-augmented objective:
\begin{equation}
    \mathcal{L}_{\text{local}} = \sum_{\ell=1}^L \sum_{j=1}^N g_{j}^{(\ell)}\,\mathcal{L}_{\text{CTC}}^{(\ell,j)}    
\end{equation}

where $\mathcal{L}_{\text{CTC}}^{(\ell,j)}$ denotes the CTC loss from expert $j$ in layer $\ell$, and $g_{j}^{(\ell)}$ is its routing probability, analogous to the intermediate CTC objective (Eq.~\ref{eq:interctc}).  
By minimizing this loss, the router is encouraged to assign higher weights to experts that yield lower CTC loss, thereby improving routing quality and enhancing overall recognition performance.  

The final training objective combines the global CTC loss, the accent-aware auxiliary loss, and the expert-level CTC supervision:
\begin{equation}
\mathcal{L} = \mathcal{L}_{\text{CTC}} 
+ \beta \, \mathcal{L}_{\text{local}}
+ \gamma \, \mathcal{L}_{\text{accent}}, 
\end{equation}
where $\beta$ and $\gamma$ are weighting coefficients.  

In this way, expert-level CTC supervision directly links routing to transcription quality, encouraging experts to specialize in improving ASR accuracy rather than only modeling accent distinctions.

\subsection{Accent-Agnostic Training}
\label{sec:accentagnostictraining}

To stabilize routing and improve generalization, we adopt a two-stage training strategy.  
The first stage focuses on specialization: both \textsc{Accent-Moe} and \textsc{Moe-Ctc} are trained with accent-aware routing (Figure~\ref{fig:moectc}, Section~\ref{sec:accent-aware}), where the router is guided by an explicit biasing term and an auxiliary accent classification loss to encourage expert alignment with specific accents.  

The second stage promotes generalization by removing these accent-specific signals, allowing the router to autonomously select experts.  
In this phase, \textsc{Accent-Moe} is optimized solely with the global CTC objective, while \textsc{Moe-Ctc} continues to leverage both global and expert-level CTC supervision ($\mathcal{L}_{\text{local}}$).  
This staged design enables the model to first learn accent-specific expertise and then generalize to unseen accents by aligning expert selection with recognition quality.

\begin{table*}[h!]
\resizebox{\textwidth}{!}{%
\begin{tabular}{ccc|cccccc|cccccccccc}
\hline
\multicolumn{3}{c|}{Configuration} & \multicolumn{6}{c|}{Seen Accent} & \multicolumn{10}{c}{Unseen Accent} \\ \hline
\multicolumn{1}{c|}{\begin{tabular}[c]{@{}c@{}}encoder \\ size\end{tabular}} &
\multicolumn{1}{c|}{model} &
\begin{tabular}[c]{@{}c@{}}parameter\\ size\end{tabular} &
AUS & CAN & UK & SCT & \multicolumn{1}{c|}{US} &
\begin{tabular}[c]{@{}c@{}}Seen \\ average\end{tabular} &
AFR & HKG & IND & IRL & MAL & NWZ & PHL & SGP &
\multicolumn{1}{c|}{WLS} & \begin{tabular}[c]{@{}c@{}}Unseen\\ average\end{tabular} \\ \hline

\multicolumn{1}{c|}{} & \multicolumn{1}{c|}{FastConformer} & 12.78M &
11.6 & 15.5 & 15.1 & 11.4 & \multicolumn{1}{c|}{13.1} & 13.3 &
20.0 & 27.4 & 29.5 & 22.5 & 30.8 & 18.0 & 27.1 & 34.1 &
\multicolumn{1}{c|}{17.1} & 25.2 \\ \cdashline{2-3}
\multicolumn{1}{c|}{} & \multicolumn{1}{c|}{Inter-CTC} & 12.78M &
11.3 & 14.3 & 15.1 & 11.0 & \multicolumn{1}{c|}{13.2} & 13.0 &
19.4 & 26.6 & 29.3 & 21.4 & 30.1 & 18.1 & 26.9 & 33.8 &
\multicolumn{1}{c|}{16.9} & 24.7 \\ \cdashline{2-3}
\multicolumn{1}{c|}{} & \multicolumn{1}{c|}{MoE} & 13.72M &
9.8 & 12.7 & 12.9 & 8.6 & \multicolumn{1}{c|}{11.5} & 11.1 &
16.4 & 24.4 & 25.3 & 19.5 & 26.5 & 15.3 & 23.2 & 31.2 &
\multicolumn{1}{c|}{16.2} & 22.0 \\ \cdashline{2-3}
\multicolumn{1}{c|}{} & \multicolumn{1}{c|}{\begin{tabular}[c]{@{}c@{}}\textsc{Accent}\\ \textsc{Moe}\end{tabular}} & 13.72M &
8.5 & 11.2 & 12.2 & 7.5 & \multicolumn{1}{c|}{10.8} & 10.0 &
16.0 & 23.8 & 24.6 & 18.9 & 26.8 & 14.8 & 22.9 & 30.5 &
\multicolumn{1}{c|}{15.8} & 21.6 \\ \cdashline{2-3}
\multicolumn{1}{c|}{\multirow{-5}{*}{Small}} & \multicolumn{1}{c|}{\textsc{Moe-Ctc}} & 16.62M &
\cellcolor[HTML]{CFE2F3}8.2 & \cellcolor[HTML]{CFE2F3}10.5 & \cellcolor[HTML]{CFE2F3}11.4 & \cellcolor[HTML]{CFE2F3}6.6 &
\multicolumn{1}{c|}{\cellcolor[HTML]{CFE2F3}10.1} & \cellcolor[HTML]{CFE2F3}9.4 &
\cellcolor[HTML]{CFE2F3}15.7 & \cellcolor[HTML]{CFE2F3}22.8 & \cellcolor[HTML]{CFE2F3}23.2 & \cellcolor[HTML]{CFE2F3}18.3 &
\cellcolor[HTML]{CFE2F3}25.5 & \cellcolor[HTML]{CFE2F3}13.7 & \cellcolor[HTML]{CFE2F3}22.5 & \cellcolor[HTML]{CFE2F3}29.6 &
\multicolumn{1}{c|}{\cellcolor[HTML]{CFE2F3}15.2} & \cellcolor[HTML]{CFE2F3}20.7 \\ \hline

\multicolumn{1}{c|}{} & \multicolumn{1}{c|}{FastConformer} & 26.39M &
7.3 & 11.4 & 10.3 & 5.7 & \multicolumn{1}{c|}{10.0} & 8.9 &
15.4 & 22.6 & 22.8 & 16.5 & 24.3 & 12.8 & 21.4 & 28.6 &
\multicolumn{1}{c|}{14.1} & 19.8 \\ \cdashline{2-3}
\multicolumn{1}{c|}{} & \multicolumn{1}{c|}{Inter-CTC} & 26.39M &
7.3 & 11.2 & 10.4 & 5.1 & \multicolumn{1}{c|}{9.8} & 8.8 &
15.5 & 22.8 & 22.9 & 16.1 & 23.8 & 12.2 & 21.1 & 27.3 &
\multicolumn{1}{c|}{13.7} & 19.5 \\ \cdashline{2-3}
\multicolumn{1}{c|}{} & \multicolumn{1}{c|}{\textsc{Moe}} & 28.37M &
7.0 & 10.3 & 9.7 & 5.0 & \multicolumn{1}{c|}{9.9} & 8.4 &
15.5 & 22.4 & 22.1 & 16.2 & 23.6 & 11.9 & 20.4 & 26.4 &
\multicolumn{1}{c|}{12.9} & 19.0 \\ \cdashline{2-3}
\multicolumn{1}{c|}{} & \multicolumn{1}{c|}{\begin{tabular}[c]{@{}c@{}}\textsc{Accent}\\ \textsc{Moe}\end{tabular}} & 28.37M &
6.3 & 9.3 & 8.5 & 4.7 & \multicolumn{1}{c|}{8.0} & 7.4 &
13.8 & 21.3 & 20.0 & 15.5 & 21.3 & 11.1 & 19.9 & 24.9 &
\multicolumn{1}{c|}{10.7} & 17.6 \\ \cdashline{2-3}
\multicolumn{1}{c|}{\multirow{-5}{*}{Medium}} & \multicolumn{1}{c|}{\textsc{Moe-Ctc}} & 32.58M &
\cellcolor[HTML]{CFE2F3}5.9 & \cellcolor[HTML]{CFE2F3}8.7 & \cellcolor[HTML]{CFE2F3}8.4 & \cellcolor[HTML]{CFE2F3}4.3 &
\multicolumn{1}{c|}{\cellcolor[HTML]{CFE2F3}7.2} & \cellcolor[HTML]{CFE2F3}6.9 &
\cellcolor[HTML]{CFE2F3}12.9 & \cellcolor[HTML]{CFE2F3}19.4 & \cellcolor[HTML]{CFE2F3}18.3 & \cellcolor[HTML]{CFE2F3}14.8 &
\cellcolor[HTML]{CFE2F3}21.1 & \cellcolor[HTML]{CFE2F3}10.7 & \cellcolor[HTML]{CFE2F3}18.9 & \cellcolor[HTML]{CFE2F3}24.2 &
\multicolumn{1}{c|}{\cellcolor[HTML]{CFE2F3}7.9} & \cellcolor[HTML]{CFE2F3}16.5 \\ \hline

\multicolumn{1}{c|}{} & \multicolumn{1}{c|}{FastConformer} & 115.60M &
5.5 & 9.4 & 7.8 & 3.9 & \multicolumn{1}{c|}{7.7} & 6.9 &
13.3 & 19.8 & 18.5 & 15.6 & 23.6 & 10.8 & 19.2 & 24.2 &
\multicolumn{1}{c|}{10.4} & 17.3 \\ \cdashline{2-3}
\multicolumn{1}{c|}{} & \multicolumn{1}{c|}{Inter-CTC} & 115.60M &
5.6 & 9.2 & 7.9 & 4.1 & \multicolumn{1}{c|}{7.5} & 6.9 &
12.9 & 19.5 & 17.9 & 15.3 & 23.8 & 10.5 & 18.3 & 23.3 &
\multicolumn{1}{c|}{9.9} & 16.8 \\ \cdashline{2-3}
\multicolumn{1}{c|}{} & \multicolumn{1}{c|}{\textsc{Moe}} & 123.48M &
5.3 & 8.8 & 7.3 & 4.0 & \multicolumn{1}{c|}{6.8} & 6.4 &
12.5 & 19.9 & 17.8 & 14.5 & 19.5 & 8.8 & 17.4 & 20.6 &
\multicolumn{1}{c|}{7.5} & 15.4 \\ \cdashline{2-3}
\multicolumn{1}{c|}{} & \multicolumn{1}{c|}{\begin{tabular}[c]{@{}c@{}}\textsc{Accent}\\ \textsc{Moe}\end{tabular}} & 123.48M &
4.9 & 7.9 & 6.8 & 3.6 & \multicolumn{1}{c|}{6.1} & 5.9 &
12.4 & 16.3 & 17.0 & 13.0 & 18.3 & 8.2 & 15.2 & 20.2 &
\multicolumn{1}{c|}{7.0} & 14.2 \\ \cdashline{2-3}
\multicolumn{1}{c|}{\multirow{-5}{*}{Large}} & \multicolumn{1}{c|}{\textsc{Moe-Ctc}} & 131.90M &
\cellcolor[HTML]{CFE2F3}4.4 & \cellcolor[HTML]{CFE2F3}7.8 & \cellcolor[HTML]{CFE2F3}6.1 & \cellcolor[HTML]{CFE2F3}3.2 &
\multicolumn{1}{c|}{\cellcolor[HTML]{CFE2F3}5.9} & \cellcolor[HTML]{CFE2F3}5.5 &
\cellcolor[HTML]{CFE2F3}11.4 & \cellcolor[HTML]{CFE2F3}12.2 & \cellcolor[HTML]{CFE2F3}16.8 & \cellcolor[HTML]{CFE2F3}12.6 &
\cellcolor[HTML]{CFE2F3}15.1 & \cellcolor[HTML]{CFE2F3}7.7 & \cellcolor[HTML]{CFE2F3}12.6 & \cellcolor[HTML]{CFE2F3}17.9 &
\multicolumn{1}{c|}{\cellcolor[HTML]{CFE2F3}6.3} & \cellcolor[HTML]{CFE2F3}12.5 \\ \hline

\end{tabular}%
}
\caption{WER (\%) on the \textsc{MCV-Accent-Test} set, comprising 5 seen and 9 unseen accents.  
All models are pretrained on \textsc{LibriSpeech-960h} and subsequently fine-tuned on \textsc{MCV-Accent-100h}.  
Highlighted cells indicate the best WER within each encoder size group.}

\label{tab:main}
\end{table*}

\section{Experimental Setup}

\subsection{Datasets}

We conduct experiments on the \textbf{\textsc{Mcv-Accent}} benchmark \cite{codebook2023}, 
which is derived from the English portion of the Mozilla CommonVoice corpus \cite{commonvoice}.  
CommonVoice is a large-scale crowd sourced dataset covering diverse speakers and accents worldwide.  
The \textsc{MCV-Accent} split provides two training subsets (100h and 600h), 
along with corresponding development and test sets.  
Summary statistics are presented in Table~\ref{tab:mcv_accent} and Table~\ref{tab:mcv_accent_devtest} in the Appendix.  

The training and development sets include five \emph{seen} accents—Australia, Canada, England, Scotland, and the United States—while the test set introduces nine additional \emph{unseen} accents to evaluate cross-accent generalization.  

To simulate realistic training strategies, all models are first pretrained on 960 hours of English speech dataset \textbf{\textsc{LibriSpeech-960h}} \cite{librispeech} 
and then fine-tuned on \textsc{MCV-Accent}, 
following a common pipeline where large-scale standard English pretraining 
enhances downstream recognition of accented speech.

\subsection{Models}

All experiments are implemented using the NeMo framework \cite{nemo} 
and trained on eight NVIDIA A100 GPUs (80GB each).  

\paragraph{Baseline Models.}
We first build baseline FastConformer encoders of three sizes—\emph{Small}, \emph{Medium}, and \emph{Large}—as summarized in Table~\ref{tab:conformer}.  
To ensure fair comparison with our MoE-based architectures, we also include intermediate FastConformer variants with approximately \textbf{46M} and \textbf{76M} parameters, matching the parameter scale of the MoE extensions.

\paragraph{MoE-Based Models.}
On top of the baseline Conformers, we additionally build three MoE variants: a standard \textbf{{MoE}} without accent supervision, an accent-aware \textbf{\textsc{Accent-Moe}}, and an expert-level supervised \textbf{\textsc{Moe-Ctc}} (Table~\ref{tab:moectc}).

Each MoE model contains $L=3$ MoE layers inserted at the 4th, 8th, and 12th encoder blocks, 
with $N=5$ experts per layer matching the number of \emph{seen accents}.  
Unless otherwise specified, all MoE variants share the same default settings:
$K=2$ for Top-K expert selection,
accent prior strength $\alpha=2$,  
local CTC loss coefficient 
$\beta = \tfrac{1}{2 \cdot (L \times N)}$,  
and accent classification loss weight $\gamma=0.1$.  

\paragraph{Tokenizer and Decoding.}
We employ a 1024-token BPE vocabulary trained on \textsc{MCV-Accent-100h}.  
For decoding, we apply greedy CTC decoding without any external language model or beam search.

\subsection{Training Configuration}
All models are trained with a global batch size of 1024 using mixed-precision (FP16) training.
During both \textsc{LibriSpeech-960h} pretraining and \textsc{MCV-Accent} fine-tuning, checkpoints are saved at the end of each epoch, and the best model is selected based on the lowest validation WER, with a maximum of 500 epochs.
We employ the AdamW optimizer with an initial learning rate of $1\times10^{-4}$ for pretraining and $1\times10^{-5}$ for fine-tuning, using a cosine annealing schedule with warmup.

For \textsc{Accent-MoE} and \textsc{Moe-Ctc}, 
we use a two-stage training strategy (Section~\ref{sec:accentagnostictraining}).  
Models are first trained with accent-aware routing, and the best checkpoint from this stage 
is then fine-tuned for 20 additional epochs without accent supervision.  
The final model is selected based on the best validation WER from this second stage.

\section{Results}
\label{sec:results}

\subsection{Main Result}

Table~\ref{tab:main} presents the Word Error Rate (WER) of all models across seen and unseen accents under various encoder sizes 
(Table~\ref{tab:conformer},~\ref{tab:moectc}).  
Across all settings, the proposed \textsc{Moe-Ctc} achieves the lowest WER, 
demonstrating clear improvements over the FastConformer baselines. 
With the \textbf{\emph{Small}} encoder, it yields relative WER reductions (WERR) of 
\textbf{29.3\%} on seen and \textbf{17.3\%} on unseen accents, 
and with the \textbf{\emph{Large}} encoder, \textbf{20.3\%} and \textbf{27.8\%}, respectively.  
Notably, the gains on unseen accents increase with model capacity, suggesting that our approach generalizes more effectively as encoder size grows.

To further examine the effect of intermediate supervision, 
we introduce an \emph{Inter-CTC} variant that adds intermediate CTC losses to the FastConformer baseline.  
This variant yields consistent yet modest improvements, confirming that intermediate supervision stabilizes training and enhances recognition quality.

Beyond this baseline, we compare the \textsc{MoE} and \textsc{Accent-MoE} variants, 
which share identical parameter sizes but differ in whether accent-aware supervision is applied.  
Across small, medium, and large encoders, 
\textsc{Accent-MoE} achieves average WERRs of 9.9\%, 11.9\%, and 7.8\% on seen accents, 
and 1.8\%, 7.4\%, and 7.8\% on unseen accents, respectively—highlighting the consistent advantage of accent-aware routing.

Building on these results, \textsc{Moe-Ctc} demonstrates the most robust and generalizable performance, 
outperforming \textsc{Accent-MoE} by an additional \textbf{6.5\%} WERR on seen and \textbf{7.5\%} on unseen accents across all encoder sizes.  
These findings suggest that expert-level CTC supervision enables \textsc{Moe-Ctc} to further enhance recognition accuracy across both seen and unseen accents.

To isolate the effect of model capacity, we control for parameter size in subsequent analyses.  
For fairness, we also compare \textsc{Moe-Ctc} and FastConformer under equal parameter budgets, 
along with prior \textsc{Mcv-Accent} benchmark results, as detailed in the following subsection.

\begin{table}[t]
\resizebox{\columnwidth}{!}{%
\begin{tabular}{l|c|c|c}
\hline
model & parameter & \begin{tabular}[c]{@{}c@{}}Seen\\ ALL\end{tabular} & \begin{tabular}[c]{@{}c@{}}Unseen\\ ALL\end{tabular} \\ \hline\hline
Conformer \cite{conformer} $\dagger$ & 43M & 14.0 & 23.7 \\ \cline{1-2}
MTL \cite{zhang2021e2e} $\dagger$ & 43M & 14.1 & 23.7 \\ \cline{1-2}
DAT \cite{das2021best} $\dagger$ & 43M & 14.0 & 23.4 \\ \cline{1-2}
\citet{codebook2023} $\dagger$ & 46M & 13.6 & 22.9 \\ \hline\hline
FastConformer & 46.89M & 13.8 & 23.7 \\ \cline{1-2}
\textsc{Moe-Ctc} & 46.91M & \cellcolor[HTML]{CFE2F3}\textbf{12.7} & \cellcolor[HTML]{CFE2F3}\textbf{22.3} \\ \hline
\end{tabular}%
}
\caption{Weighted average WER (\%) on the \textsc{Mcv-Accent-Test} set. All models are trained on the \textbf{\textsc{Mcv-Accent-100h}} training split. $\dagger$ indicates results reported by \citet{codebook2023}.}
\label{tab:46M}
\end{table}

\subsection{Benchmark}

To compare with prior studies on \textsc{Mcv-Accent}, we evaluate two benchmark settings.  
The first uses a 46M-parameter model trained solely on the \textsc{Mcv-Accent-100h} split 
without Librispeech pretraining, following \citet{codebook2023} 
(as shown in Table~\ref{tab:46M}).  
The second employs a 76M-parameter model trained on the larger \textsc{Mcv-Accent-600h} split, 
also without pretraining, following \citet{codebook2024} (Table~\ref{tab:76M}).  

The \textit{Seen-ALL} and \textit{Unseen-ALL} scores are calculated as weighted averages 
based on the number of samples per accent (Table~\ref{tab:mcv_accent_devtest}), 
which differs from the unweighted averages reported in Table~\ref{tab:main}.

As shown in Table~\ref{tab:46M}, removing \textsc{Librispeech-960h} pretraining leads to higher WER across all models.
Our proposed \textsc{Moe-Ctc} still outperforms other baselines and \citet{codebook2023}, 
although the improvement is relatively modest, achieving a WERR of \textbf{6.6\%} on seen and \textbf{2.6\%} on unseen accents.  
When compared to our FastConformer baseline, the gains are slightly larger 
(8.0\% on seen and 5.9\% on unseen), but remain smaller than those observed in Table~\ref{tab:main}.  
This indicates that training with only 100 hours of data is insufficient to fully exploit the advantages of the proposed method.

In contrast, utilizing the larger \textsc{Mcv-Accent-600h} training set yields a substantial performance improvement.  
As shown in Table~\ref{tab:76M}, \textsc{Moe-Ctc} outperforms previous best model \cite{codebook2024} with larger WERR margins of \textbf{18.4\%} on seen and \textbf{5.9\%} on unseen accents.  







\begin{table}[t]
\resizebox{\columnwidth}{!}{%
\begin{tabular}{l|c|c|c}
\hline
model & parameter & \begin{tabular}[c]{@{}c@{}}Seen\\ ALL\end{tabular} & \begin{tabular}[c]{@{}c@{}}Unseen\\ ALL\end{tabular} \\ \hline \hline
HuBERT \cite{hsu2021hubert} $\dagger$ & 74M & 3.87 & 9.49 \\ \cline{1-2}
MTL \cite{zhang2021e2e} $\dagger$ & 74M & 3.76 & 9.37 \\ \cline{1-2}
DAT \cite{das2021best} $\dagger$ & 74M & 3.83 & 9.30 \\ \cline{1-2}
\citet{codebook2024} $\dagger$ & 76M & 3.80 & 9.19 \\ \hline \hline
FastConformer & 76.70M & 3.53 & 9.52 \\ \cline{1-2}
\textsc{Moe-Ctc} & 76.26M & \cellcolor[HTML]{CFE2F3}\textbf{3.10} & \cellcolor[HTML]{CFE2F3}\textbf{8.65} \\ \hline
\end{tabular}%
}
\caption{Weighted average WER(\%) on the \textsc{Mcv-Accent-Test} set. All models are trained on the \textbf{\textsc{Mcv-Accent-600h}} training split. 
$\dagger$ indicates results reported by \citet{codebook2024}.}

\label{tab:76M}
\end{table}

\section{Analysis}
\label{sec:analysis}

\subsection{Effect of Accent-Agnostic Training}
\label{sec:agnostic-training}
We further investigate the impact of accent-agnostic training (Section~\ref{sec:accentagnostictraining}) for both \textsc{Accent-Moe} and \textsc{Moe-Ctc} under the \emph{Large} encoder configuration. 
Table~\ref{tab:2stage} presents the results across different training stages.  

For \textsc{Accent-Moe}, the accent-agnostic only training corresponds to the standard MoE baseline reported in Table~\ref{tab:main}. 
Introducing an initial accent-aware stage yields consistent gains on both seen and unseen accents 
(6.4 $\rightarrow$ 6.1 and 15.4 $\rightarrow$ 15.1, respectively). 
When followed by accent-agnostic fine-tuning, the model achieves further improvements, 
reducing WER to \textbf{5.9} (↓ 3.8\%) on seen and \textbf{14.2} (↓ 6.0\%) on unseen accents. 
These results confirm that accent-agnostic training helps the router generalize beyond explicit accent supervision.

A similar trend is observed in \textsc{Moe-Ctc}, where incorporating accent-agnostic fine-tuning after accent-aware pretraining 
leads to a more pronounced improvement—achieving \textbf{5.5} (↓ 5.2\%) on seen and \textbf{12.5} (↓ 12.0\%) on unseen accents. 
This indicates that our expert-level CTC supervision is especially effective when combined with accent-agnostic training, 
allowing the model to retain accent sensitivity while improving generalization to unseen accents.

\begin{table}[t!]
\centering
\resizebox{\linewidth}{!}{
\begin{tabular}{lcll}
\toprule
Model & Training Stage & Seen Avg. & Unseen Avg. \\
\midrule \midrule
MoE & Accent-agnostic only & 6.4 & 15.4 \\ \midrule
\textsc{Accent-Moe} & Accent-aware & 6.1 & 15.1 \\ 
  & +Accent-agnostic & 
5.9 {\small \textcolor{ForestGreen}{($\downarrow$ \textbf{3.8}\%)}} &  
14.2 {\small \textcolor{ForestGreen}{($\downarrow$ \textbf{6.0}\%)}} \\ \midrule
\textsc{Moe-Ctc} & Accent-aware & 5.8 & 14.2 \\ 
  & +Accent-agnostic & 
5.5 {\small \textcolor{ForestGreen}{($\downarrow$ \textbf{5.2}\%)}} & 
12.5 {\small \textcolor{ForestGreen}{($\downarrow$ \textbf{12.0}\%)}} \\
\bottomrule
\end{tabular}}
\caption{Average WER(\%) for seen and unseen accents under different training stages. 
Green text indicates WERR compared to the accent-aware stage.}
\label{tab:2stage}

\end{table}



\subsection{Accent Routing Performance}
While \citet{zuluaga2023commonaccent} report 95\% accuracy on a 16-way accent classification task using w2v2-XLSR (315M) trained on the CommonAccent dataset, their model is optimized solely for accent identification. 
In contrast, our router is jointly trained with ASR, making the task considerably more challenging. 
Under this multitask setting, the 76M \textsc{Moe-Ctc} model (in Table~\ref{tab:76M}), trained on 600 hours of data, attains 72\% top-1 routing accuracy on the final MoE layer before accent-agnostic training. 
A similar observation was made by \citet{codebook2023}, who found that auxiliary accent classifiers for expert selection offered limited benefit over inference-time beam search, suggesting that perfect accent discrimination is neither achievable nor necessary within ASR training. 

\begin{figure}[h!]
    \centering
    \includegraphics[width=0.85\linewidth]{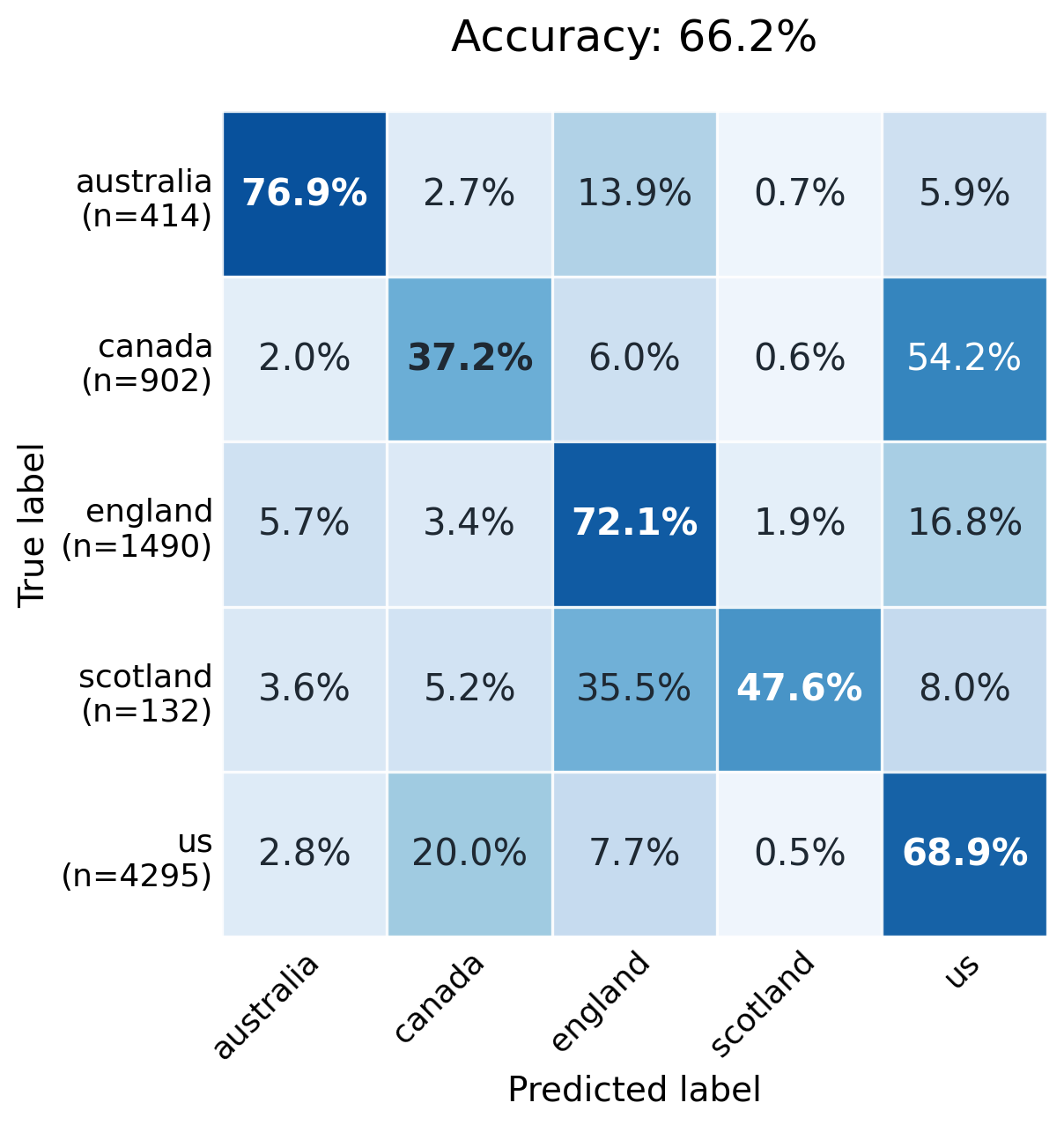}
\caption{Matrix of router gating probabilities ($g_{i,j}$) at the final (\textit{3rd}) \textsc{Moe-Ctc} module of the 76M model trained on \textsc{Mcv-Accent-600h}. Each row represents an accent, and the probabilities in each row sum to $\sum_{j=1}^{N=5} g_{i,j} = 1$.}

    \label{fig:confusion}
\end{figure}

After accent-agnostic fine-tuning, the router in \textsc{Moe-Ctc} still preserves meaningful accent organization, achieving 66.2\% top-1 accuracy (Figure~\ref{fig:confusion}). 
The confusion matrix reveals that Canadian speech is often routed to the U.S. expert and Scottish to the England expert, indicating that routing aligns with acoustically and geographically correlated accents even without explicit supervision. 
This adaptive redistribution promotes better generalization by guiding expert selection toward recognition-oriented specialization (Section~\ref{sec:agnostic-training}).


\subsection{Oracle Accent Routing Setting}

To estimate the upper bound of our architecture under ideal conditions, 
we conduct an oracle experiment where ground-truth accent labels are provided at inference time. 
Specifically, we enforce hard expert routing by applying the biasing term 
(Eq.~\ref{eq:bias}) with maximum strength ($\alpha = \infty$), ensuring that each 
accent exclusively activates its designated expert. 
As shown in Table~\ref{tab:golden}, providing oracle labels yields the lowest WER of 
\textbf{5.2\%} under accent-aware training and \textbf{5.3\%} when combined with accent-agnostic fine-tuning, 
compared to \textbf{5.8\%} and \textbf{5.5\%} without label access. 
These results confirm that oracle routing achieves the best performance by leveraging accent-specific specialization, 
while accent-agnostic training slightly reduces specialization but improves generalization for label-free inference.

\begin{table}[t]
\resizebox{\columnwidth}{!}{%
\begin{tabular}{cccc}
\midrule
model & \begin{tabular}[c]{@{}c@{}}Training Stage\end{tabular} & \begin{tabular}[c]{@{}c@{}}Accent Label\end{tabular} & \begin{tabular}[c]{@{}c@{}}Seen Average\end{tabular} \\ \midrule \midrule
MoE-CTC & Accent-aware & No & 5.8 (5.82) \\ \midrule
MoE-CTC & Both & No & 5.5 (5.48) \\ \midrule\midrule 
MoE-CTC & Accent-aware & Yes & \textbf{5.2} (5.20) \\ \midrule 
MoE-CTC & Both & Yes & 5.3 (5.32) \\ \midrule
\end{tabular}%
}
\caption{WER(\%) under the oracle accent routing setting with \emph{Large} encoder. 
Parentheses denote values reported to two decimal places.}

\label{tab:golden}
\end{table}



\section{Conclusion}
We presented \textsc{Moe-Ctc}, a Mixture-of-Experts architecture enhanced with intermediate CTC supervision for accent-robust ASR. 
By combining expert-level CTC heads with a two-stage transition from accent-aware to accent-agnostic training, 
the model achieves relative WER reductions of up to 29.3\% and 27.8\% on seen and unseen accents, respectively, 
on the \textsc{Mcv-Accent} benchmark. 
These results highlight that aligning expert routing with transcription quality substantially improves both robustness and generalization in accented speech recognition.



\section*{Limitations}
  
While \textsc{Moe-Ctc} demonstrates substantial improvements in accented speech recognition, its sequence-level routing assumes discrete accent boundaries, which may not generalize to mixed or code-switched speech. 
The model also relies on accent labels during early training, limiting applicability in unsupervised settings. 
Although sparsely activated, the additional MoE modules increase training cost and latency. 
Finally, experiments are limited to English accents on \textsc{Mcv-Accent}; broader multilingual evaluation and analysis of expert interpretability remain future work.

\bibliography{custom}

@inproceedings{ctc,
  title={Connectionist temporal classification: labelling unsegmented sequence data with recurrent neural networks},
  author={Graves, Alex and Fern{\'a}ndez, Santiago and Gomez, Faustino and Schmidhuber, J{\"u}rgen},
  booktitle={Proceedings of the 23rd international conference on Machine learning},
  pages={369--376},
  year={2006}
}

@inproceedings{interctc,
  title={Intermediate loss regularization for ctc-based speech recognition},
  author={Lee, Jaesong and Watanabe, Shinji},
  booktitle={ICASSP 2021-2021 IEEE International Conference on Acoustics, Speech and Signal Processing (ICASSP)},
  pages={6224--6228},
  year={2021},
  organization={IEEE}
}

@inproceedings{scctc,
  title={Relaxing the Conditional Independence Assumption of CTC-Based ASR by Conditioning on Intermediate Predictions},
  author={Nozaki, Jumon and Komatsu, Tatsuya},
  booktitle={Proc. Interspeech 2021},
  pages={3735--3739},
  year={2021}
}

@inproceedings{komatsu2022better,
  author    = {Komatsu, T. and Fujita, Y. and Lee, J. and Lee, L. and Watanabe, S. and Kida, Y.},
  title     = {Better Intermediates Improve {CTC} Inference},
  booktitle = {Proceedings of Interspeech 2022},
  pages     = {4965--4969},
  year      = {2022},
  doi       = {10.21437/Interspeech.2022-11276}
}

@inproceedings{hojo2024boosting,
  author    = {Hojo, K. and Wakabayashi, Y. and Ohta, K. and Ogawa, A. and Kitaoka, N.},
  title     = {Boosting {CTC}-based {ASR} using Inter-layer Attention-based {CTC} Loss},
  booktitle = {Proceedings of Interspeech 2024},
  pages     = {2860--2864},
  year      = {2024},
  doi       = {10.21437/Interspeech.2024-1776}
}

@inproceedings{codebook2023,
    title = "Accented Speech Recognition With Accent-specific Codebooks",
    author = "Prabhu, Darshan  and
      Jyothi, Preethi  and
      Ganapathy, Sriram  and
      Unni, Vinit",
    editor = "Bouamor, Houda  and
      Pino, Juan  and
      Bali, Kalika",
    booktitle = "Proceedings of the 2023 Conference on Empirical Methods in Natural Language Processing",
    month = dec,
    year = "2023",
    address = "Singapore",
    publisher = "Association for Computational Linguistics",
    url = "https://aclanthology.org/2023.emnlp-main.444/",
    doi = "10.18653/v1/2023.emnlp-main.444",
    pages = "7175--7188",
}

@inproceedings{codebook2024,
  author    = {Prabhu, D. and Gupta, A. and Nitsure, O. and Jyothi, P. and Ganapathy, S.},
  title     = {Improving Self-supervised Pre-training using Accent-Specific Codebooks},
  booktitle = {Proceedings of Interspeech 2024},
  pages     = {2310--2314},
  year      = {2024},
  doi       = {10.21437/Interspeech.2024-2438}
}

@inproceedings{radford2023robust,
  title={Robust speech recognition via large-scale weak supervision},
  author={Radford, Alec and Kim, Jong Wook and Xu, Tao and Brockman, Greg and McLeavey, Christine and Sutskever, Ilya},
  booktitle={Proceedings of the 40th International Conference on Machine Learning},
  pages={28492--28518},
  year={2023}
}

@article{baevski2020wav2vec,
  title={wav2vec 2.0: A Framework for Self-Supervised Learning of Speech Representations},
  author={Baevski, Alexei and Zhou, Yuhao and Mohamed, Abdelrahman and Auli, Michael},
  journal={Advances in Neural Information Processing Systems},
  volume={33},
  pages={12449--12460},
  year={2020}
}

@article{hsu2021hubert,
  title={Hubert: Self-supervised speech representation learning by masked prediction of hidden units},
  author={Hsu, Wei-Ning and Bolte, Benjamin and Tsai, Yao-Hung Hubert and Lakhotia, Kushal and Salakhutdinov, Ruslan and Mohamed, Abdelrahman},
  journal={IEEE/ACM transactions on audio, speech, and language processing},
  volume={29},
  pages={3451--3460},
  year={2021},
  publisher={IEEE}
}

@inproceedings{babu2022xls,
  title={XLS-R: Self-supervised Cross-lingual Speech Representation Learning at Scale},
  author={Babu, Arun and Wang, Changhan and Tjandra, Andros and Lakhotia, Kushal and Xu, Qiantong and Goyal, Naman and Singh, Kritika and von Platen, Patrick and Saraf, Yatharth and Pino, Juan and others},
  booktitle={Proc. Interspeech 2022},
  pages={2278--2282},
  year={2022}
}

@inproceedings{conformer,
  author    = {Gulati, A. and Qin, J. and Chiu, C.-C. and Parmar, N. and Zhang, Y. and Yu, J. and Han, W. and Wang, S. and Zhang, Z. and Wu, Y. and Pang, R.},
  title     = {Conformer: Convolution-augmented Transformer for Speech Recognition},
  booktitle = {Proceedings of Interspeech 2020},
  pages     = {5036--5040},
  year      = {2020},
  doi       = {10.21437/Interspeech.2020-3015}
}

@inproceedings{fastconformer,
  title={Fast conformer with linearly scalable attention for efficient speech recognition},
  author={Rekesh, Dima and Koluguri, Nithin Rao and Kriman, Samuel and Majumdar, Somshubra and Noroozi, Vahid and Huang, He and Hrinchuk, Oleksii and Puvvada, Krishna and Kumar, Ankur and Balam, Jagadeesh and others},
  booktitle={2023 IEEE Automatic Speech Recognition and Understanding Workshop (ASRU)},
  pages={1--8},
  year={2023},
  organization={IEEE}
}

@article{nemo,
  title={Nemo: a toolkit for building ai applications using neural modules},
  author={Kuchaiev, Oleksii and Li, Jason and Nguyen, Huyen and Hrinchuk, Oleksii and Leary, Ryan and Ginsburg, Boris and Kriman, Samuel and Beliaev, Stanislav and Lavrukhin, Vitaly and Cook, Jack and others},
  journal={arXiv preprint arXiv:1909.09577},
  year={2019}
}

@inproceedings{bagat2025mixture,
  author    = {Bagat, R. and Illina, I. and Vincent, E.},
  title     = {Mixture of {LoRA} Experts for Low-Resourced Multi-Accent Automatic Speech Recognition},
  booktitle = {Proceedings of Interspeech 2025},
  pages     = {1143--1147},
  year      = {2025},
  doi       = {10.21437/Interspeech.2025-775}
}

@inproceedings{zhou2024dialectmoe,
  title={DialectMoE: An End-to-End Multi-Dialect Speech Recognition Model with Mixture-of-Experts},
  author={Jie, Zhou and Shengxiang, Gao and Zhengtao, Yu and Ling, Dong and Wenjun, Wang},
  booktitle={Proceedings of the 23rd Chinese National Conference on Computational Linguistics (Volume 1: Main Conference)},
  pages={1148--1159},
  year={2024}
}

@inproceedings{hu2023mixture,
  title={Mixture-of-Expert Conformer for Streaming Multilingual ASR},
  author={Hu, Ke and Li, Bo and Sainath, Tara and Zhang, Yu and Beaufays, Fran{\c{c}}oise},
  booktitle={Proc. Interspeech 2023},
  pages={3327--3331},
  year={2023}
}

@inproceedings{das2021best,
  title={Best of Both Worlds: Robust Accented Speech Recognition with Adversarial Transfer Learning},
  author={Das, Nilaksh and Bodapati, Sravan and Sunkara, Monica and Srinivasan, Sundararajan and Chau, Duen Horng},
  booktitle={Proc. Interspeech 2021},
  pages={1314--1318},
  year={2021}
}

@inproceedings{zhang2021e2e,
  title={E2E-Based Multi-Task Learning Approach to Joint Speech and Accent Recognition},
  author={Zhang, Jicheng and Peng, Yizhou and Pham, Van Tung and Xu, Haihua and Huang, Hao and Chng, Eng Siong},
  booktitle={Proc. Interspeech 2021},
  pages={1519--1523},
  year={2021}
}

@inproceedings{commonvoice,
    title = "Common Voice: A Massively-Multilingual Speech Corpus",
    author = "Ardila, Rosana  and
      Branson, Megan  and
      Davis, Kelly  and
      Kohler, Michael  and
      Meyer, Josh  and
      Henretty, Michael  and
      Morais, Reuben  and
      Saunders, Lindsay  and
      Tyers, Francis  and
      Weber, Gregor",
    editor = "Calzolari, Nicoletta  and
      B{\'e}chet, Fr{\'e}d{\'e}ric  and
      Blache, Philippe  and
      Choukri, Khalid  and
      Cieri, Christopher  and
      Declerck, Thierry  and
      Goggi, Sara  and
      Isahara, Hitoshi  and
      Maegaard, Bente  and
      Mariani, Joseph  and
      Mazo, H{\'e}l{\`e}ne  and
      Moreno, Asuncion  and
      Odijk, Jan  and
      Piperidis, Stelios",
    booktitle = "Proceedings of the Twelfth Language Resources and Evaluation Conference",
    month = may,
    year = "2020",
    address = "Marseille, France",
    publisher = "European Language Resources Association",
    url = "https://aclanthology.org/2020.lrec-1.520/",
    pages = "4218--4222",
    language = "eng",
    ISBN = "979-10-95546-34-4",
}

@inproceedings{zuluaga2023commonaccent,
  title={CommonAccent: Exploring Large Acoustic Pretrained Models for Accent Classification Based on Common Voice},
  author={Zuluaga-Gomez, Juan and Ahmed, Sara and Visockas, Danielius and Subakan, Cem},
  booktitle={Proc. Interspeech 2023},
  pages={5291--5295},
  year={2023}
}

@inproceedings{l2arctic,
  author    = {Zhao, G. and Sonsaat, S. and Silpachai, A. and Lucic, I. and Chukharev-Hudilainen, E. and Levis, J. and Gutierrez-Osuna, R.},
  title     = {L2-{ARCTIC}: A Non-native English Speech Corpus},
  booktitle = {Proceedings of Interspeech 2018},
  pages     = {2783--2787},
  year      = {2018},
  doi       = {10.21437/Interspeech.2018-1110}
}

@inproceedings{globe,
  title={GLOBE: A High-quality English Corpus with Global Accents for Zero-shot Speaker Adaptive Text-to-Speech},
  author={Wang, Wenbin and Song, Yang and Jha, Sanjay},
  booktitle={Proc. Interspeech 2024},
  pages={1365--1369},
  year={2024}
}

@inproceedings{shor2019personalizing,
  title={Personalizing ASR for Dysarthric and Accented Speech with Limited Data},
  author={Shor, Joel and Emanuel, Dotan and Lang, Oran and Tuval, Omry and Brenner, Michael and Cattiau, Julie and Vieira, Fernando and McNally, Maeve and Charbonneau, Taylor and Nollstadt, Melissa and others},
  booktitle={Proc. Interspeech 2019},
  pages={784--788},
  year={2019}
}

@INPROCEEDINGS{librispeech,
  author={Panayotov, Vassil and Chen, Guoguo and Povey, Daniel and Khudanpur, Sanjeev},
  booktitle={2015 IEEE International Conference on Acoustics, Speech and Signal Processing (ICASSP)}, 
  title={Librispeech: An ASR corpus based on public domain audio books}, 
  year={2015},
  volume={},
  number={},
  pages={5206-5210},
  doi={10.1109/ICASSP.2015.7178964}}

@article{qian2022layer,
  title={Layer-Wise Fast Adaptation for End-to-End Multi-Accent Speech Recognition},
  author={Qian, Yanmin and Gong, Xun and Huang, Houjun},
  journal={IEEE/ACM TRANSACTIONS ON AUDIO, SPEECH, AND LANGUAGE PROCESSING},
  volume={30},
  year={2022}
}

@inproceedings{tomanek-etal-2021-residual,
    title = "Residual Adapters for Parameter-Efficient {ASR} Adaptation to Atypical and Accented Speech",
    author = "Tomanek, Katrin  and
      Zayats, Vicky  and
      Padfield, Dirk  and
      Vaillancourt, Kara  and
      Biadsy, Fadi",
    editor = "Moens, Marie-Francine  and
      Huang, Xuanjing  and
      Specia, Lucia  and
      Yih, Scott Wen-tau",
    booktitle = "Proceedings of the 2021 Conference on Empirical Methods in Natural Language Processing",
    month = nov,
    year = "2021",
    address = "Online and Punta Cana, Dominican Republic",
    publisher = "Association for Computational Linguistics",
    url = "https://aclanthology.org/2021.emnlp-main.541/",
    doi = "10.18653/v1/2021.emnlp-main.541",
    pages = "6751--6760"
}

@inproceedings{sun2018dat,
  title={Domain adversarial training for accented speech recognition},
  author={Sun, Sining and Yeh, Ching-Feng and Hwang, Mei-Yuh and Ostendorf, Mari and Xie, Lei},
  booktitle={ICASSP},
  pages={4854--4858},
  year={2018}
}

@inproceedings{das2021adversarial,
  title={Best of both worlds: Robust accented speech recognition with adversarial transfer learning},
  author={Das, Nilaksh and Bodapati, Sravan and Sunkara, Monica and Srinivasan, Sundararajan and Chau, Duen Horng},
  booktitle={INTERSPEECH},
  year={2021}
}

@inproceedings{hu2020reDAT,
  title={ReDAT: Accent-invariant representation for end-to-end ASR by domain adversarial training with relabeling},
  author={Hu, Hu and Yang, Xuesong and Raeesy, Zeynab and Guo, Jinxi and Keskin, Gokce and Arsikere, Harish and Rastrow, Ariya and Stolcke, Andreas and Maas, Roland},
  booktitle={INTERSPEECH},
  year={2020}
}

@inproceedings{unni2020coupled,
  title={Coupled training of sequence-to-sequence models for accented speech recognition},
  author={Unni, Vinit and Joshi, Nitish and Jyothi, Preethi},
  booktitle={ICASSP},
  pages={8254--8258},
  year={2020}
}

@inproceedings{jain2018accentemb,
  title={Improved accented speech recognition using accent embeddings and multi-task learning},
  author={Jain, Abhinav and Upreti, Minali and Jyothi, Preethi},
  booktitle={Interspeech},
  pages={2454--2458},
  year={2018}
}

@inproceedings{rao2017hierarchical,
  title={Multi-accent speech recognition with hierarchical grapheme based models},
  author={Rao, Kanishka and Sak, Ha{\c{s}}im},
  booktitle={ICASSP},
  pages={4815--4819},
  year={2017}
}

@article{winata2020fastadapt,
  title={Learning fast adaptation on cross-accented speech recognition},
  author={Winata, Genta Indra and Cahyawijaya, Samuel and Liu, Zihan and Lin, Zhaojiang and Madotto, Andrea and Xu, Peng and Fung, Pascale},
  journal={arXiv preprint arXiv:2003.01901},
  year={2020}
}

@inproceedings{chen2015ivector,
  title={Improving deep neural networks based multi-accent Mandarin speech recognition using i-vectors and accent-specific top layer},
  author={Chen, Mingming and Yang, Zhanlei and Liang, Jizhong and Li, Yanpeng and Liu, Wenju},
  booktitle={INTERSPEECH},
  year={2015}
}

@inproceedings{li2017multidialect,
  title={Multi-dialect speech recognition with a single sequence-to-sequence model},
  author={Li, Bo and Sainath, Tara N and Sim, Khe Chai and Bacchiani, Michiel and Weinstein, Eugene and Nguyen, Patrick and Chen, Zhifeng and Wu, Yonghui and Rao, Kanishka},
  booktitle={ICASSP},
  pages={4749--4753},
  year={2017}
}

@inproceedings{viglino2019fusion,
  title={End-to-end accented speech recognition},
  author={Viglino, Thibault and Motl{\'\i}{\v{c}}ek, Petr and Cernak, Milos},
  booktitle={INTERSPEECH},
  year={2019}
}

@article{graham2024whisper_accents,
  title = {Evaluating OpenAI's Whisper ASR: Performance analysis across diverse accents},
  author = {Graham, Calbert and Roll, Nathan},
  journal = {JASA Express Letters},
  year = {2024},
  doi = {10.1121/10.0024876}
}

@article{olatunji2023afrispeech200,
  title = {AfriSpeech-200: Pan-African Accented Speech Dataset for ASR},
  author = {Olatunji, T. and Afonja, T. and Dossou, B. F. P. and others},
  journal = {Transactions of the Association for Computational Linguistics (TACL)},
  year = {2023}
}

@inproceedings{Dheram2022toward,
  title={Toward Fairness in Speech Recognition: Discovery and mitigation of performance disparities},
  author={DHERAM, PRANAV and Ramakrishnan, Murugesan and Raju, Anirudh and Chen, I-Fan and King, Brian and Powell, Katherine and Saboowala, Melissa and Shetty, Karan and Stolcke, Andreas},
  booktitle={Proc. Interspeech 2022},
  pages={1268--1272},
  year={2022}
}

@inproceedings{do2024improving,
  title={Improving Accented Speech Recognition Using Data Augmentation Based on Unsupervised Text-to-Speech Synthesis},
  author={Do, Cong-Thanh and Imai, Shuhei and Doddipatla, Rama and Hain, Thomas},
  booktitle={2024 32nd European Signal Processing Conference (EUSIPCO)},
  pages={136--140},
  year={2024},
  organization={IEEE}
}

@misc{li2021accentrobustautomaticspeechrecognition,
      title={Accent-Robust Automatic Speech Recognition Using Supervised and Unsupervised Wav2vec Embeddings}, 
      author={Jialu Li and Vimal Manohar and Pooja Chitkara and Andros Tjandra and Michael Picheny and Frank Zhang and Xiaohui Zhang and Yatharth Saraf},
      year={2021},
      eprint={2110.03520},
      archivePrefix={arXiv},
      primaryClass={eess.AS},
      url={https://arxiv.org/abs/2110.03520}, 
}

@inproceedings{you2021speechmoe,
  title={SpeechMoE: Scaling to Large Acoustic Models with Dynamic Routing Mixture of Experts},
  author={You, Zhao and Feng, Shulin and Su, Dan and Yu, Dong},
  booktitle={Proc. Interspeech 2021},
  pages={2077--2081},
  year={2021}
}

@article{fedus2022switch,
  title={Switch transformers: Scaling to trillion parameter models with simple and efficient sparsity},
  author={Fedus, William and Zoph, Barret and Shazeer, Noam},
  journal={Journal of Machine Learning Research},
  volume={23},
  number={120},
  pages={1--39},
  year={2022}
}

@article{shazeer2017outrageously,
  title={Outrageously large neural networks: The sparsely-gated mixture-of-experts layer},
  author={Shazeer, Noam and Mirhoseini, Azalia and Maziarz, Krzysztof and Davis, Andy and Le, Quoc and Hinton, Geoffrey and Dean, Jeff},
  journal={arXiv preprint arXiv:1701.06538},
  year={2017}
}

\appendix

\section{Appendix}
\label{sec:appendix}

\subsection{\textsc{Mcv-Accent} Dataset}

For English text normalization, we adopt the whisper-normalizer\footnote{\url{https://kurianbenoy.github.io/whisper_normalizer/}} toolkit to ensure consistent text preprocessing across both the training (Table~\ref{tab:mcv_accent}) and evaluation (Table~\ref{tab:mcv_accent_devtest}) sets.

\begin{table}[h!]
\centering
\resizebox{\linewidth}{!}{%
\begin{tabular}{l|rrr|rrr}
\toprule
\multirow{2}{*}{Accent} & \multicolumn{3}{c|}{\textbf{\textsc{Mcv-Accent-100h}}} & \multicolumn{3}{c}{\textbf{\textsc{Mcv-Accent-600h}}} \\
\cmidrule(lr){2-4} \cmidrule(lr){5-7}
 & Count & Duration (h) &  & Count & Duration (h) &  \\
\midrule
US        & 46,385  & 64.08 & & 289,177 & 399.88 & \\
England   & 14,590  & 19.51 & & 90,081  & 119.93 & \\
Australia & 5,018   & 6.95  & & 31,355  & 43.56  & \\
Canada    & 4,815   & 6.79  & & 29,436  & 41.13  & \\
Scotland  & 1,669   & 2.69  & & 10,123  & 16.21  & \\
\bottomrule
\end{tabular}
}
\caption{Comparison of accent distribution and duration in \textsc{Mcv-Accent-100h} and \textsc{Mcv-Accent-600h} subsets from \citet{codebook2023}.}
\label{tab:mcv_accent}
\end{table}

\begin{table}[h!]
\centering
\resizebox{\linewidth}{!}{%
\begin{tabular}{l|rr|rr}
\toprule
\multirow{2}{*}{Accent} & \multicolumn{2}{c|}{\textbf{\textsc{Mcv-Accent-dev}}} & \multicolumn{2}{c}{\textbf{\textsc{Mcv-Accent-test}}} \\
\cmidrule(lr){2-3} \cmidrule(lr){4-5}
 & Count & Duration (h) & Count & Duration (h) \\
\midrule
US          & 7,050 & 8.32 & 4,295 & 4.87 \\
England     & 2,775 & 3.22 & 1,490 & 1.65 \\
Australia   & 3,004 & 4.33 &   414 & 0.46 \\
Canada      &   966 & 1.16 &   902 & 1.21 \\
Scotland    &   233 & 0.23 &   132 & 0.16 \\
New Zealand & –     & –    & 1,620 & 2.11 \\
Ireland     & –     & –    & 1,423 & 1.94 \\
African     & –     & –    & 1,213 & 1.71 \\
Philippines & –     & –    &   631 & 0.90 \\
Indian      & –     & –    &   492 & 0.58 \\
Singapore   & –     & –    &   477 & 0.64 \\
Hongkong    & –     & –    &   409 & 0.52 \\
Malaysia    & –     & –    &   262 & 0.39 \\
Wales       & –     & –    &   219 & 0.27 \\
\bottomrule
\end{tabular}
}
\caption{Statistics of validation (\textsc{Mcv-Accent-dev}) and test (\textsc{Mcv-Accent-test}) sets.}
\label{tab:mcv_accent_devtest}
\end{table}

\subsection{Model Configuration Details}

Table~\ref{tab:conformer} summarizes the configurations of the FastConformer baselines, including the total number of parameters, encoder layers, hidden dimension, and multi-head attention heads. 
The corresponding configurations for \textsc{Moe-Ctc} are presented in Table~\ref{tab:moectc}. 
The \emph{Small}, \emph{Medium}, and \emph{Large} variants share identical FastConformer settings as in Table~\ref{tab:conformer}, with additional MoE modules inserted between encoder blocks. 
For the 46M and 76M variants, the base FastConformer encoders are slightly downsized—by adjusting the number of layers and hidden dimensions—to ensure comparable total parameter counts after adding the MoE layers.

\begin{table}[h!]
\centering
\resizebox{\columnwidth}{!}{%
\begin{tabular}{lcccc}
\toprule
\textbf{Name} & \textbf{Params} & \textbf{Layers} & \textbf{d\_model} & \textbf{Heads} \\
\midrule
Small   & 12.78M  & 16 & 176 & 4 \\
Medium  & 26.39M  & 16 & 256 & 4 \\
46M     & 46.89M  & 18 & 324 & 4 \\
76M     & 76.70M  & 18 & 416 & 4 \\
Large   & 115.60M & 18 & 512 & 8 \\
\bottomrule
\end{tabular}
}
\caption{FastConformer configurations with parameter size, depth, hidden dimension ($d_{\text{model}}$), and multi-head attention heads.}
\label{tab:conformer}
\end{table}

\begin{table}[h!]
\centering
\resizebox{\columnwidth}{!}{%
\begin{tabular}{lcccc}
\toprule
\textbf{Name} & \textbf{Params} & \textbf{Layers} & \textbf{d\_model} & \textbf{Heads} \\
\midrule
Small   & 16.62M (12.78M)   & 16 & 176 & 4 \\
Medium  & 32.58M (26.39M)   & 16 & 256 & 4 \\
46M     & \textbf{46.91M} (38.85M)   & \textbf{16} & \textbf{312} & 4 \\
76M     & \textbf{76.26M} (65.51M)   & \textbf{18} & \textbf{384} & 4 \\
Large   & 123.48M (115.60M) & 18 & 512 & 8 \\
\bottomrule
\end{tabular}
}
\caption{\textsc{Moe-Ctc} model configurations. Notation follows Table~\ref{tab:conformer}. Values in parentheses indicate the base FastConformer parameters, excluding \textsc{Moe-Ctc} modules.}
\label{tab:moectc}
\end{table}

\newpage

\subsection{Adding Spare Experts}
By default, we set the number of experts to $5$, matching the number of seen accents for accent-aware routing.
In the \emph{Large} configuration, we further increase the capacity by adding three additional experts, resulting in a total of $8$ experts.
During training, these extra experts are randomly assigned to approximately 20\% of the training samples from all seen accents, encouraging exposure to diverse acoustic patterns without explicit bias.
As shown in Table~\ref{tab:my-table}, increasing the number of experts consistently improves recognition performance.
For \textsc{Accent-MoE}, expanding from 5 to 8 experts reduces WER from 5.86 to 5.83 (\textbf{0.5\%} relative) on seen accents and from 14.18 to 14.12 (\textbf{0.4\%}) on unseen accents.
The improvement is more pronounced for \textsc{Moe-Ctc}, achieving a WER reduction from 5.48 to 5.39 (\textbf{1.6\%}) on seen and from 12.51 to 12.28 (\textbf{1.9\%}) on unseen accents.
These consistent gains indicate that the additional experts act as global experts, learning accent-invariant representations that complement accent-specific specialization and enhance overall generalization.

\begin{table}[t]
\centering
\resizebox{\columnwidth}{!}{%
\begin{tabular}{cccll}
\toprule
Model & \begin{tabular}[c]{@{}c@{}}Number of \\ Experts\end{tabular} & 
\begin{tabular}[c]{@{}c@{}}Parameter \\ size\end{tabular} & 
\begin{tabular}[c]{@{}c@{}}Seen \\ average\end{tabular} & 
\begin{tabular}[c]{@{}c@{}}Unseen \\ average\end{tabular} \\ 
\midrule \midrule
\textsc{Accent-Moe} & 5 & 123.48M & 5.86 & 14.18 \\ 
\textsc{Accent-Moe} & 8 & 129.89M & 
\textbf{5.83}~\textcolor{green!50!black}{\small($\downarrow$ 0.5\%)} & 
14.12~\textcolor{green!50!black}{\small($\downarrow$ 0.4\%)} \\ 
\midrule \midrule
\textsc{Moe-Ctc} & 5 & 131.90M & 5.48 & 12.51 \\ 
\textsc{Moe-Ctc} & 8 & 141.37M & 
\textbf{5.39}~\textcolor{green!50!black}{\small($\downarrow$ 1.6\%)} & 
\textbf{12.28}~\textcolor{green!50!black}{\small($\downarrow$ 1.9\%)} \\ 
\bottomrule
\end{tabular}%
}
\caption{WER(\%) comparison between 5- and 8-expert variants. 
Green text denotes relative WER reduction (WERR) of 8-expert over 5-expert models.}
\label{tab:my-table}
\end{table}



\subsection{\textsc{Ctc-Head} sharing}

\textsc{Moe-Ctc} (Figure~\ref{fig:moectc}) equips each expert with its own \textsc{Ctc-Head}, enabling expert-specific alignment but also increasing the overall parameter count.
To examine more lightweight alternatives, we explore several \textsc{Ctc-Head} sharing strategies.
As summarized in Table~\ref{tab:head}, \emph{Full separation} denotes the default configuration where each expert has an independent head.
In the \emph{Layer-wise sharing} setup, a single \textsc{Ctc-Head} is shared among all experts within the same MoE layer, while \emph{Global sharing} removes intermediate heads entirely, using only the final global CTC head for all experts.
Layer-wise sharing slightly degrades WER on both seen (+2.2\%) and unseen (+4.5\%) accents, yet it remains competitive and far superior to \textsc{Accent-Moe} while reducing parameters by about 5\%.
In contrast, global sharing results in a more substantial degradation (+8.0\% on seen, +16.1\% on unseen), even worse than \textsc{Accent-Moe} with no expert-specific head.
We hypothesize that each MoE layer produces representations of varying transcription quality—earlier layers generate coarse alignments, while deeper layers refine them—thus separate heads at each layer help preserve alignment fidelity and stabilize expert specialization.

\begin{table}[t!]
\centering
\resizebox{\columnwidth}{!}{%
\begin{tabular}{clcll}
\toprule
Model & \begin{tabular}[c]{@{}l@{}}CTC-Head\\ Sharing\end{tabular} & 
\begin{tabular}[c]{@{}c@{}}Parameter \\ size\end{tabular} & 
\begin{tabular}[c]{@{}c@{}}Seen \\ average\end{tabular} & 
\begin{tabular}[c]{@{}c@{}}Unseen \\ average\end{tabular} \\ 
\midrule
\multirow{3}{*}{\textsc{Moe-Ctc}} 
 & Full separation & 131.90M & \textbf{5.48} & \textbf{12.51} \\ 
 & Layer-wise sharing & 125.59M & 5.60~\textcolor{red!60!black}{\small($\uparrow$2.2\%)} & 13.07~\textcolor{red!60!black}{\small($\uparrow$4.5\%)} \\ 
 & Global sharing & 124.01M & 5.92~\textcolor{red!60!black}{\small($\uparrow$8.0\%)} & 14.52~\textcolor{red!60!black}{\small($\uparrow$16.1\%)} \\ 
\midrule
\textsc{Accent-Moe} & No Head & 123.48M & 5.86 & 14.18 \\ 
\bottomrule
\end{tabular}%
}
\caption{WER(\%) comparison of different CTC head sharing strategies in \textsc{Moe-Ctc}. 
Red text indicates relative WER degradation compared to the full separation baseline.}
\label{tab:head}
\end{table}

\end{document}